\documentclass[runningheads]{llncs}
\usepackage[title]{appendix}
\usepackage[T1]{fontenc}
\usepackage[dvipsnames]{xcolor}
\usepackage{multirow}

\usepackage{graphicx}
\usepackage{makecell}
\usepackage{amsmath}
\usepackage[labelfont=bf]{caption}
\usepackage{subcaption}
\usepackage{tablefootnote}
\usepackage{algorithm2e}
\usepackage{array,multirow,graphicx}
 \usepackage{float}
 \usepackage{amsfonts}
 
\RestyleAlgo{ruled}
\usepackage{xspace}

\begin{document}

\title{Leveraging Symbolic Regression for Heuristic Design in the Traveling Thief Problem}

\author{Andrew Ni\inst{1}\orcidID{0000-0003-3480-4536}\and 
Lee Spector\inst{2,3}\orcidID{10000-0001-5299-4797} }

\authorrunning{Ni, et al.}
\institute{Amherst College, Amherst MA 01002, USA \email{hickory62133@gmail.com}\and
Amherst College, Amherst, MA 01002, USA\and
University of Massachusetts Amherst, Amherst, MA 01003, USA \email{lspector@amherst.edu}}


\maketitle    

\begin{abstract}
The Traveling Thief Problem is an NP-hard combination of the well known traveling salesman and knapsack packing problems. In this paper, we use symbolic regression to learn useful features of near-optimal packing plans, which we then use to design efficient metaheuristic genetic algorithms for the traveling thief algorithm. By using symbolic regression again to initialize the metaheuristic GA with near-optimal individuals, we are able to design a fast, interpretable, and effective packing initialization scheme. Comparisons against previous initialization schemes validates our algorithm design. 
\end{abstract}
\section{Introduction}

The Traveling Thief Problem (TTP) is an NP-hard combination of the Traveling Salesman (TSP) and 0-1 Knapsack Packing (KP) problems designed to be more representative of real life situations\cite{ttpbenchmarks,ttpinitial}. In a TTP problem, a thief is presented with a set of $n$ cities $\text{Cities}=\{i,\cdots,n\}$ and a distance function $d(i,j):\mathbb{Z}^2\rightarrow\mathbb{Z}$, where each city $i$ except the first one has $m$ items with weight $w_{ik}$ and profit $p_{ik}$. The thief has to visit every city in the given set, starting and ending at city 1. At each city, they may put any of the available items into their knapsack, which has capacity $W$. However, the more full their knapsack is, the slower they travel, and a renting cost $R$ is incurred for each time unit spent traveling. Therefore, for a given tour $[c_1,\cdots c_n,c_1]$ and packing plan $\text{\tt packed}(i,k):\{1...n\}\times\{1...m\}\rightarrow\{0,1\}$, where 0 denotes an unpacked item and 1 denotes a packed item, their final objective value will be $$\sum_{i=1}^{n}\sum_{k=1}^m \text{\tt packed}(i,k)*p_{ik}-R\times\left(\frac{d(n,1)}{v_{\text{max}}-\nu W_n}+\sum_{i=1}^{n-1}\frac{d(i,i+1)}{v_{\text{max}}-\nu W_{i}}\right)$$ where $v_{\text{max}}$ and $v_{\text{min}}$ are the maximum and minimum travel speeds, respectively, $\nu=\frac{v_{\text{max}}-v_{\text{min}}}{W}$ is the decrease in speed per unit weight, and \nopagebreak $W_i=\sum_{i'=0}^i\sum_{k=1}^m\text{\tt packed}(i',k)*w_{ik}$ is the total weight of the knapsack at the $i^{\rm th}$ city.

In this paper, we use machine learning to design packing heuristics for the traveling thief problem that achieve a better "fit" than previous human-designed heuristics. We first show that there is a smooth, regular relationship between the standardized item profitability ratio (IPR) and the standardized distance to the end of the tour (rDist) of items packed in high-quality packing plans. We then use symbolic regression to discover combinations of the two features IPR and rDist that are important to the abovementioned relationship. Identification of promising feature combinations then gives rise to a family of metaheuristic genetic algorithms, in which each individual is roughly a vector of the coefficients for these combinations and defines a boundary line separating packed items from unpacked items. By again utilizing symbolic regression to analyze the optimal individuals produced from our metaheuristic genetic algorithm, we are able to directly predict the parameter values of the optimal individual to high precision, greatly accelerating the GA and stabilizing its performance. In this way, we obtain a family of effective and high-performing packing initialization heuristics for the traveling thief problem. We perform experiments to choose the best initialization scheme from this family, and then demonstrate the superior speed and problem-solving performance of our heuristic compared to the previous SOTA.

Our paper is organized as follows: In Section 2, we give a brief overview of related work, including previous TTP solvers, and especially previous initialization heuristics. In Section 3 we gradually infer our initialization heuristic by characterizing optimal packing solutions to the traveling thief problem. Finally, in Section 4, we describe our experiments and results, comparing our heuristics both among themselves and against previous initialization heuristics. Empirical results show that our heuristics are a better "fit" for TTP instances, achieving higher objective values in fewer objective value evaluations than previous methods.

\section{Background and Related Work}

\subsection{TTP Benchmark Suite}

In the benchmark suite proposed by Polyakovskiy et al.\cite{ttpbenchmarks}, a TTP instance starts off as one of the TSP instances from the TSPLIB library\cite{tsplib,tsplibpaper}. An item factor $F$ is chosen from the set $\{1,3,5,10\}$, and $F$ items are then placed at each city other than the first one. The weights and profits of these items are determined by the knapsack problem type, which can be one of {\tt uncorr}, \newline{\tt uncorr\_similar\_weights}, and {\tt bounded\_strongly\_corr}, in order of increasing difficulty for KP-solvers\cite{combo}. In the {\tt uncorr} category, weights and profits are sampled uniformly from $[1,1000]$. In the {\tt uncorr\_similar\_weights} category, they are sampled from $[1000,1010]$ and $[1,1000]$, respectively. Finally, in the {\tt bounded\_strongly\_corr} category, the weight of the $k^{\rm th}$ item in the $i^{\rm th}$ city $w_{ik}$ is sampled uniformly from $[1,1000]$ and the profit is determined based on this weight as $p_{ik}=100+w_{ik}$. Based on the near-optimal solutions to the TSP and KP instances generated by the Chained Lin-Kernighan heuristic\cite{linkern,chainedlinkern} and the COMBO solver\cite{combo} respectively, a renting ratio is calculated as the optimal profit of the KP subproblem divided by the time taken to traverse the linkern tour while collecting the items in that optimal packing. In other words, the optimal KP packing combined with the linkern tour will give a zero TTP objective value. This prevents one subproblem dominating the other, and forces solvers to focus on both subproblems to achieve high objective values\cite{ttpbenchmarks}. Finally, the knapsack capacity is determined based on a capacity factor $C\in\{1...10\}$ as $W=\frac{C}{11}\sum_{i=1}^n\sum_{k=0}^nw_{ik}$. Altogether this gives $4\times3\times10=120$ TTP instances for each TSP instance.

\subsection{TTP Solvers}

Since the initial proposal of the traveling thief problem\cite{ttpinitial} and the creation of the TTP benchmark suite\cite{ttpbenchmarks}, many approaches for solving the problem have been developed. Polyakovskiy et al.\cite{ttpbenchmarks} generate a tour using the chained lin-kernighan heuristic\cite{linkern} and then apply either a hill-climbing local search (RLS), a (1+1) EA, or a simple heuristic (SH) to generate a good packing plan for that tour. While the EA and RLS methods perform well on smaller instances, SH takes much less time to run and performs the best on the largest TTP instances, illustrating the importance of a good initialization heuristic. 

After some time, a large-scale benchmark was conducted of all proposed TTP solvers, and a new algorithm was proposed, which attempts to choose the best previously proposed solver based on information about the specific TTP instance. Besides the aforementioned RLS, EA, and SH algorithms, benchmarked algorithms included DH\cite{dh}, MATLS\cite{matls}, S1-S5 and C1-C5\cite{packiterative}, CS2SA\cite{cs2sa}, and M3/M3B and M4/M4B\cite{maxminants}.

The density-based heuristic DH and the CoSolver algorithm were proposed by Bonyandi et al.\cite{dh}. In the DH heuristic, similarly to SH, items are assigned scores based on the change in objective value obtained when going from a zero packing plan to one which only picks that item. Items are then picked in decreasing order of their scores until no further improvement in objective value can be obtained. On the other hand, CoSolver attempts to decompose the TTP problem into multiple subproblems, interleaving tour improvement for a given packing plan with packing improvement for a given tour. These algorithms were then improved to give the solvers CS2SA\cite{ma2b-cs2sa}, an improvement to CoSolver with 2-Opt and Simulated Annealing, MA2B\cite{ma2b-cs2sa} a population-based memetic algorithm with MPX-based crossover\cite{mpx}, and CS2SA$^*$ and CS2SA-R\cite{cs2sa*}, two improvements to the CS2SA algorithm.

The Memetic Algorithm with Two-Stage Local Search\cite{matls} follows a similar approach, alternating between tour and packing improvement. They also develop a novel packing initialization heuristic {\tt Insertion}, using three different approximations to the change in objective value upon packing an item. The first approximation is the same as that used in DH. The second approximation is a worst-case approximation assuming that the knapsack is already full. Finally, their third approximation is an expected-value approximation assuming that the knapsack weight is distributed evenly along the tour. 

Faulkner et al.\cite{packiterative} propose several algorithms mainly based on their new heuristic {\tt packIterative}. In {\tt packIterative}, items are give scores based on their profit, weight, and remaining distance as $\text{score}_{ik}=\frac{p_{ik}^\alpha}{w_{ik}^\alpha d_i}$, where the parameter $\alpha$ is optimized in an outer loop based on a binary-search-like algorithm. Items are collected in decreasing order of their scores until no further objective increase is possible. However, to decrease the number of costly objective value computations, they propose to pack some fraction of the total items at each step, which in practice is generally 1\%\footnote{https://cs.adelaide.edu.au/~optlog/research/combinatorial.php}\footnote{github.com/majid75/CoCo} If at any point the objective value decreases, they revert their changes and halve the step size. This proceeds until no improvement is possible or the step size reaches 1. 

In a different direction, Wagner et al.\cite{maxminants} pursue swarm-intelligence-based TTP solvers with their M3, M3B, M4, and M4B solvers. For a given packing plan, they use an ant colony optimization algorithm to improve the tour. While their algorithm does not scale well to large instances, it finds very good solutions on smaller instances. 

Finally, Majid et al.\cite{coco} have recently proposed a cooperative coordination heuristic PGCH for simultaneously improving the tour and the packing plan, as well as a directed local search algorithm MBFS for quickly improving packing plans. They show that this approach yields better objective values than previous SOTA algorithms, obtaining record objective values in most large problem instances studied.

Many of these previous approaches often start and restart by generating one or more decent individuals by generating a good tour, either using the lin-kernighan heuristic\cite{linkern,chainedlinkern} or something faster but less optimal like the quick-boruvka algorithm\cite{chainedlinkern}, and then generating an initial packing for that individual using a human-designed heuristic like {\tt packIterative}\cite{packiterative} or {\tt Insertion}\cite{matls}. These heuristics generally compute a utility score for each item using a parameterized function designed to capture human intuition about which items are likely to be packed in an optimal solution. Items are then packed in decreasing order of their utility scores until no improvement in the objective value is possible.  However, these human-designed heuristics are not very flexible, as they only have zero or one learnable parameters, and if the "shape" of their score functions do not match the "shape" of high quality packing plans, then their maximum achievable objective score may be reduced. In contrast, we propose directly learning appropriate score function "shape"s from the problem instances themselves by using symbolic regression to analyze near-optimal packing plans. This gives our score functions a better "fit" to the problem at hand and results in higher objective scores.

\subsection{Symbolic Regression}

Symbolic regression\cite{koza} is a subfield of Evolutionary Computation concerned with evolving mathematical functions to fit some dataset. Due to their symbolic nature and simplicity, Symbolic regression methods are more interpretable than neural net-based models\cite{srbench}. In addition, large-scale testing on the Penn Machine Learning Benchmark (PMLB)\cite{pmlb} has shown that SR-based algorithms are competitive with if not better than other methods in terms of accuracy\cite{srbench}. Many SR methods\cite{operon,eplex,sbpgp,gplearn}, including the first system introduced by Koza\cite{koza}, are based on genetic programming (GP), evolving populations of executable data structures, commonly parse trees, that take a number of inputs and produce a numerical output. In these parse trees, the nodes are usually mathematical functions and the leaves are input features or numerical constants. Other SR approaches include methods inspired by physics\cite{alfeynman}, based on Bayesian optimization\cite{bayesian}, or utilizing reinforcement learning algorithms\cite{deepsymbolicregression}. For the sake of convenience, we reuse a previous system, the gplearn system\cite{gplearn} with DALex\cite{dalex} as the selection method, which has been shown to be fast and have good accuracy on SR problems. However, as many of the conclusions we draw are symbolic and qualitative in nature, we believe that the same results can be obtained with other SR systems as well.

\section{SR-guided Heuristic Design}

In this section we present our proposed packing initialization heuristics through a careful characterization of near-optimal packing plans. 

\subsection{Analysis of Optimal Packing Plans}

\begin{figure}
     \centering
     \includegraphics[width=\textwidth]{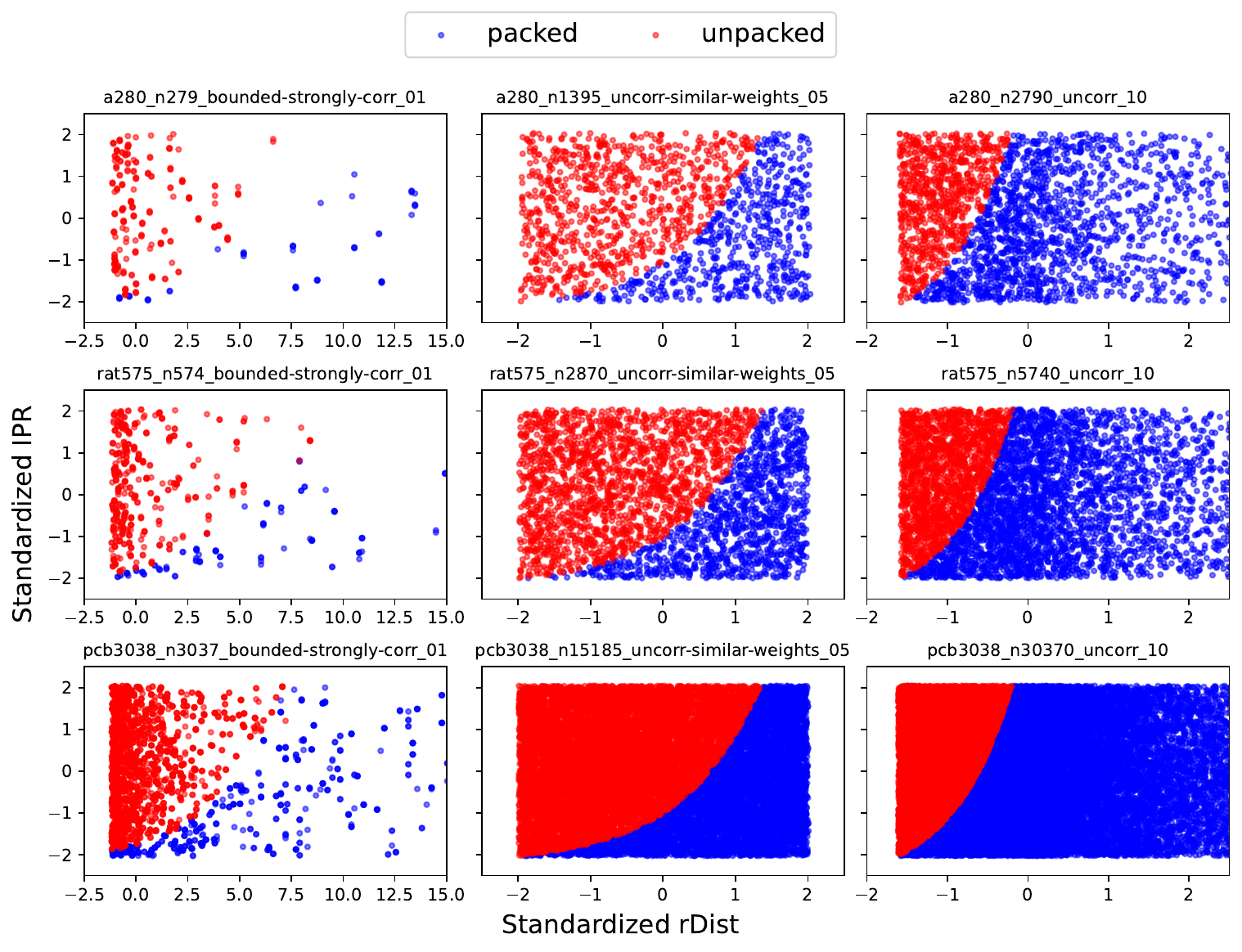}
        \caption{Plot of the packed and unpacked items in near-optimal tours generated by the (1+1) EA against the normalized item profitability ratio and distance to the end of the tour. A smooth nonlinear boundary can be seen separating packed items from unpacked items. While the specific boundary varies from instance to instance, it keeps the same overall curved shape.}
        
        \label{fig:ones_and_zeros}
\end{figure}

We first study the relationship between the standardized item profitability ratio $\text{\tt IPR}_{ik}=\frac{p_{ik}}{w_{ik}}$ and distance-to-go $\text{\tt rDist}_{ik}=d(n,1)+\sum_{i'=i}^{n-1}d(i',i'+1)$ of items collected versus uncollected in near-optimal packing plans. Due to the large ourliers in the {\tt IPR} when items have very low weights, we standardize to zero median $M$ and unit median absolute deviation $\text{\tt MAD}(\{\text{\tt IPR}_{ik}\})=\text{\tt median}(\{|\text{\tt IPR}_{ik}-M)|\})$ instead of zero mean and unit variance. In addition, as these outliers are always packed due to their light weight and high profit, we limit our attention to {\tt IPR} and {\tt rDist} values in the interval $[-2,2]$. To obtain near-optimal packing plans, we adopt the method proposed in Polyakovskiy et al.\cite{ttpbenchmarks}, generating a tour using the chained lin-kernighan heuristic\cite{linkern} and using a (1+1) hill-climbing EA to optimize the packing plan, starting from a packing plan that does not pick any items. To get high quality solutions, we run this EA for $10^6$ generations. At the end of the algorithm, we record for each item the standardized {\tt IPR} and {\tt rDist}, along with whether it was packed in the best packing plan found. Due to space limitations, we only show a few results from selected TTP instances in figure\ref{fig:ones_and_zeros}.

While combinatorial optimization problems like the traveling thief problems are inherently discrete, the plots in figure \ref{fig:ones_and_zeros} show a smooth boundary demarcating the packed items from the unpacked items. Of course, the degree of discontinuity is higher for smaller instances, so we hypothesize that the heuristics developed in this paper will work better with larger problem instances. This nicely complements the behavior of packing improvement, which can find optimal solutions to small instances very quickly but may take a long time for larger problems. 
Heuristic algorithms that assign scores to items based on their profitability ratio and remaining distance essentially draw boundary lines in this space and collect all items on one side of the line. While we do not expect the boundary line that best fits the near-optimal (1+1) EA solutions to also produce the best packing plans as an initialization heuristic, we hypothesize that the shape of the boundary line is a fundamental aspect of the traveling thief problem. Therefore, heuristics that can better match the shape of these near-optimal packing plans have a better "fit" for TTP problems, and are more likely to be able to produce packing plans with very high objectives.

To learn the shape of these boundary lines, we train a nonlinear binary classifier (NLBC) to distinguish betwen packed and unpacked items using symbolic regression. Due to computational constraints, we choose a range of smaller TTP instances with roughly log-spaced sizes: {\tt a280}, {\tt rat575}, {\tt rl1323}, and {\tt pcb3038}. From each category we take 90 problems with capacity factors $C\in\{1...10\}$, knapsack type in \{{\tt bounded\_strongly\_corr}, {\tt uncorr\_similar\_weights},{\tt uncorr}\}, and item factor $F\in\{1,5,10\}$. 

Instead of using the absolute error fitness function as in many SR systems\cite{gplearn,eplex,srbench}, we use the binary cross entropy (BCE) loss $\mathcal{L}(y_i,\hat{y_i})=-y_i\log\hat{y_i}-(1-y_i)\log(1-\hat{y_i})$ as our fitness function, where $y_i\in\{0,1\}$ is the ground-truth label and $\hat{y_i}=\frac{1}{1+e^{-x_i}}$, where $x_i$ is the output of the SR individual on the $i^{\rm th}$ training example. We use the DALex\cite{dalex} selection algorithm because it is fast and produces solutions with good accuracy. We keep the other hyperparameters unchanged from the defaults in gplearn\footnote{https://github.com/trevorstephens/gplearn} except that we reduce the number of generations to 300 and combine the results from 5 independent runs. 

At each generation, we take the individual with the best BCE loss, expand it into a sum of terms using {\tt sympy}, remove the constant coefficient from each term to get a set of features, and count the number of features. Due to computational limitations, we only apply this process to the best individual in each generation. We say that an individual dominates another individual if it has a better BCE loss and a number of terms equal to or less than that of the other individual. Throughout each run, we keep a pareto front of non-dominated best-of-generation individuals, and at the end of each run we collect the feature set of each individual in the final pareto front. We group all of the individuals collected this way from all 5 independent runs, and count the number of identical feature sets at each feature set size. To save space, we only report feature sets that appeared more than once. The results are displayed in table \ref{featuresets}.

\begin{table}
\caption{Frequency of feature sets in the pareto-fronts of evolved NLBCs, grouped by feature set size.}\label{featuresets}
\begin{tabular}{|c|c|c|c|c|c|c|c|}

\hline
    \multicolumn{2}{|c|}{size 2} & \multicolumn{2}{c|}{size 3}& \multicolumn{2}{c|}{size 4}& \multicolumn{2}{c|}{size 5} \\\hline
     $\{x_0,x_1\}$ & 108 & $\{1,x_0,x_1\}$ & 213 & $\{1,x_0,x_1,x_0x_1\}$ & 158 & $\{1,x_0,x_1,x_0x_1,x_0^2\}$ & 77\\
     $\{1,x_0\}$ & 57 & $\{x_0,x_1,x_0x_1\}$ & 10 & $\{1,x_0,x_1,x_0^2\}$ & 53& $\{1,x_0,x_1,x_0x_1,x_1^2\}$ & 46\\
     $\{1,x_1\}$ & 5 & $\{1,x_0,x_0x_1\}$ & 10 & $\{1,x_0,x_1,x_1^2\}$ & 6& $\{1,x_0,x_1,x_0^2,x_0^3\}$ & 19\\\hline
     \multicolumn{4}{|c|}{size 6} & \multicolumn{4}{c|}{size 7}  \\\hline
      \multicolumn{3}{|c|}{$\{1,x_0,x_1,x_0x_1,x_1^2,x_0x_1^2\}$} & 13 & \multicolumn{3}{c|}{$\{1,x_0,x_1,x_0x_1,x_1^2,x_0x_1^2,x_1^3\}$} & 5 \\
     \multicolumn{3}{|c|}{$\{1,x_0,x_1,x_0x_1,x_0^2,x_1^2\}$} & 11 & \multicolumn{3}{c|}{$\{1,x_0,x_1,x_0x_1,x_1^2,x_0^2x_1,x_0x_1^2\}$} & 4 \\
     \multicolumn{3}{|c|}{$\{1,x_0,x_1,x_0x_1,x_1^2,x_0^3\}$} & 10 & \multicolumn{3}{c|}{$\{1,x_0,x_1,x_0x_1,x_0^2,x_1^2,x_0x_1^2\}$} & 3 \\\hline
\end{tabular}
\end{table}

From this table, we select a few promising feature sets. In the size 3, and 4 groups, the most frequent feature set occurs much more often than the runner-up, so we choose the most frequent feature set from each group. Since the two most frequent sets in the size 5 group appear a relatively similar number of times, we choose both of them. Although the feature sets in the size 6 and 7 groups all rarely appear and are all fairly similar, we also choose the $\{1,x_0,x_1,x_0x_1,x_0^2,x_1^2\}$ feature set from that group. In this way, we identify 5 different feature sets, which we name based on the number of terms they contain: {\tt 2T} = $\{x_0,x_1\}$, {\tt 3T} = $\{1,x_0,x_1\}$, {\tt 4T} = $\{1,x_0,x_1,x_0x_1\}$, {\tt 5T0} = $\{1,x_0,x_1,x_0x_1,x_0^2\}$, {\tt 5T1} = $\{1,x_0,x_1,x_0x_1,x_1^2\}$, and {\tt 6T} = $\{1,x_0,x_1,x_0x_1,x_0^2,x_1^2\}$. 

\subsection{Analysis of Optimal Parameter Values}

Each feature set $\{f_i\}$ identified in the previous section gives rise to a metaheuristic algorithm, where individuals of the form $[w_0,\cdots,w_n]$ encode boundary lines of the form $\sum_{i=0}^nw_if_i=0$ and pack all items such that $\sum_{i=0}^nw_if_i>0$. We use a (1+1) EA, where the initial individual is the 0 vector and mutations are standard normal perturbations $\sim\mathcal{N}(0,1)$. To obtain near-optimal objective values, we conduct 4 independent runs of our EA for $10^5$ generations each. In preliminary experiments, we found that the optimal values of the $w_i$ were all relatively stable except for the "bias" term corresponding to the feature $1$, which would vary significantly depending on the quality of the tour. Since for a given TTP instance and TSP tour there is a 1-to-1 correspondence between the value of the "bias" term and the total weight packed by the plan, we replace the bias term with a "percent" term describing the packed capacity as a percentage of the knapsack capacity, which we find to be much more stable. Therefore, instead of directly encoding a boundary line, each individual is now of the form $[w_0,\cdots,w_n,p]$ and encodes a score function $\text{\tt score}=\sum_{i=0}^nw_if_i$. Items are then collected in decreasing order of their scores until reaching a total weight of $pW$, where $W$ is the knapsack capacity. 

Due to this removal of the $1$ term, the {\tt 2T} and {\tt 3T} feature sets are no longer distinct, so now our collection of feature sets becomes: {\tt 3T} = $\{x_0,x_1\}$, {\tt 4T} = $\{x_0,x_1,x_0x_1\}$, {\tt 5T0} = $\{x_0,x_1,x_0x_1,x_0^2\}$, {\tt 5T1} = $\{x_0,x_1,x_0x_1,x_1^2\}$, and \\{\tt 6T} = $\{x_0,x_1,x_0x_1,x_0^2,x_1^2\}$. Finally, as the percent term has to be in the interval $[0,1]$, we mutate it with noise sampled according to $\mathcal{N}(0,0.1)$ and clamp it to the interval $[0,1]$.

After conducting the 4 metaheuristic algorithm runs for a given feature set and a given TTP instance, we record the best individual across all runs $[w_0,\cdots,w_n,p]$. However, since scaling all the $w_i$'s of an individual by a constant does not change its performance, we first normalize the genotype so that $\sum_{i=0}^nw_i^2=1$. We then use symbolic regression again to predict these near-optimal individuals' genotype using 7 features of the TTP instance: the number of cities $x_0$, the number of items $x_1$, the renting ratio $x_2$, the knapsack capacity $x_3$, the item factor $x_4$, the maximum speed $x_5$, the minimum speed $x_6$, and the capacity factor $x_7$. Note that the names of these input variables are distinct from the names of the variables in the aformentioned feature sets, which were aliases for the standardized {\tt IPR} and {\tt rDist}. 

We use the same TTP problems as in the previous section, comprising of 90 instances each from the {\tt a280}, {\tt rat575}, {\tt rl1323}, and {\tt pcb3038} categories. However, we separate them based on their knapsack type, and conduct separate runs for problems in each knapsack type. Since our SR system is only built for single-valued outputs, we conduct a separate run for each index in the metaheuristic genotype vector. We use the same system as in the previous section, except this time we're interested in eliminating useless input variables rather than constructing useful features. Therefore, instead of using sympy to simplify best-of-generation individuals, we instead track a rough approximation of the pareto front, keeping the best individuals with program lengths less than 20, 30, and 40. We also increase the number of parallel runs to 10, which makes for a total of up to 30 distinct individuals obtained from each set of 10 runs. For each input variable, we then calculate what percentage of these individuals contain that input variable in their programs. However, since $v_{\text{max}}$ and $v_{\text{min}}$ remain the same throughout all instances in the TTP benchmark set\cite{ttpbenchmarks}, we treat them as constants and omit them from consideration. Due to space considerations, we only display the results for the {\tt 4T} feature set in table \ref{freqtable}, and put the other results in the appendix.

\begin{table}
\caption{Percent of evolved solutions containing each input variable for each parameter in the genotype of an individual from the {\tt 4T} metaheuristic GA. The highest value for each parameter-KP type combination is shown in bold. Almost all solutions contain the $x_7$ variable, whereas other variables are usually present in less than one third of the solutions.}\label{freqtable}
\begin{tabular}{|c||c|c|c|c|c||c|c|c|c|c||c|c|c|c|c|}

\hline
     & \multicolumn{5}{c||}{bounded-strongly-corr}& \multicolumn{5}{c||}{uncorr-similar-weights} &\multicolumn{5}{c|}{uncorr}\\\hline
     & $x_1$ & $x_2$ & $x_3$ & $x_4$& $x_7$&  $x_1$ & $x_2$ & $x_3$ & $x_4$&$x_7$&  $x_1$ & $x_2$ & $x_3$ & $x_4$& $x_7$ \\\hline
$w0$ & 0.17 & 0.37 & 0.07 & 0.4 & {\bf 1.0} & 0.27 & 0.1 & 0.17 & 0.17 & {\bf 1.0} & 0.2 & 0.13 & 0.07 & 0.03 & {\bf 0.97}\\
$w1$ & 0.37 & 0.17 & 0.03 & 0.43 & {\bf 1.0} & 0.1 & 0.0 & 0.0 & 0.1 & {\bf 1.0} & 0.07 & 0.0 & 0.18 & 0.0 & {\bf 1.0}\\
$w2$ & 0.1 & 0.17 & 0.03 & 0.9 & {\bf 1.0} & 0.57 & 0.5 & 0.07 & 0.6 & {\bf 1.0} & 0.4 & 0.1 & 0.07 & 0.5 & {\bf 0.93}\\
percent & 0.1 & 0.1 & 0.1 & 0.13 & {\bf 1.0} & 0.0 & 0.0 & 0.1 & 0.03 & {\bf 1.0} & 0.03 & 0.03 & 0.0 & 0.0 & {\bf 1.0}\\\hline
\end{tabular}
\end{table}

Since the capacity factor variable $x_7$ was present in almost all solutions, and other variables are only present in a few solutions, we hypothesize that the optimal parameter values depend solely on the capacity factor. To verify this hypothesis, evolved solutions using the same symbolic regression setup as before, but now restricting the set of input features to the $x_7$ variable. We then take the individual with the lowest mean absolute error, and plot it against the capacity factor, along with the rest of our dataset. We encountered two anomalous solutions, one for the $w_2$ parameter of the {\tt 5T0} feature set in the {\tt uncorr} KP type, and one for the $w_3$ parameter of the {\tt 6T} feature set. The graphs of these solutions did not match their high fitness, which we hypothesize is due to differences in handling edge conditions like dividing by zero. So, we replaced them with a linear fit. Due to limited space, we again only show the results for the {\tt 4T} feature set, and put the rest in the appendix. 

\begin{figure}
     \centering
     \includegraphics[width=\textwidth]{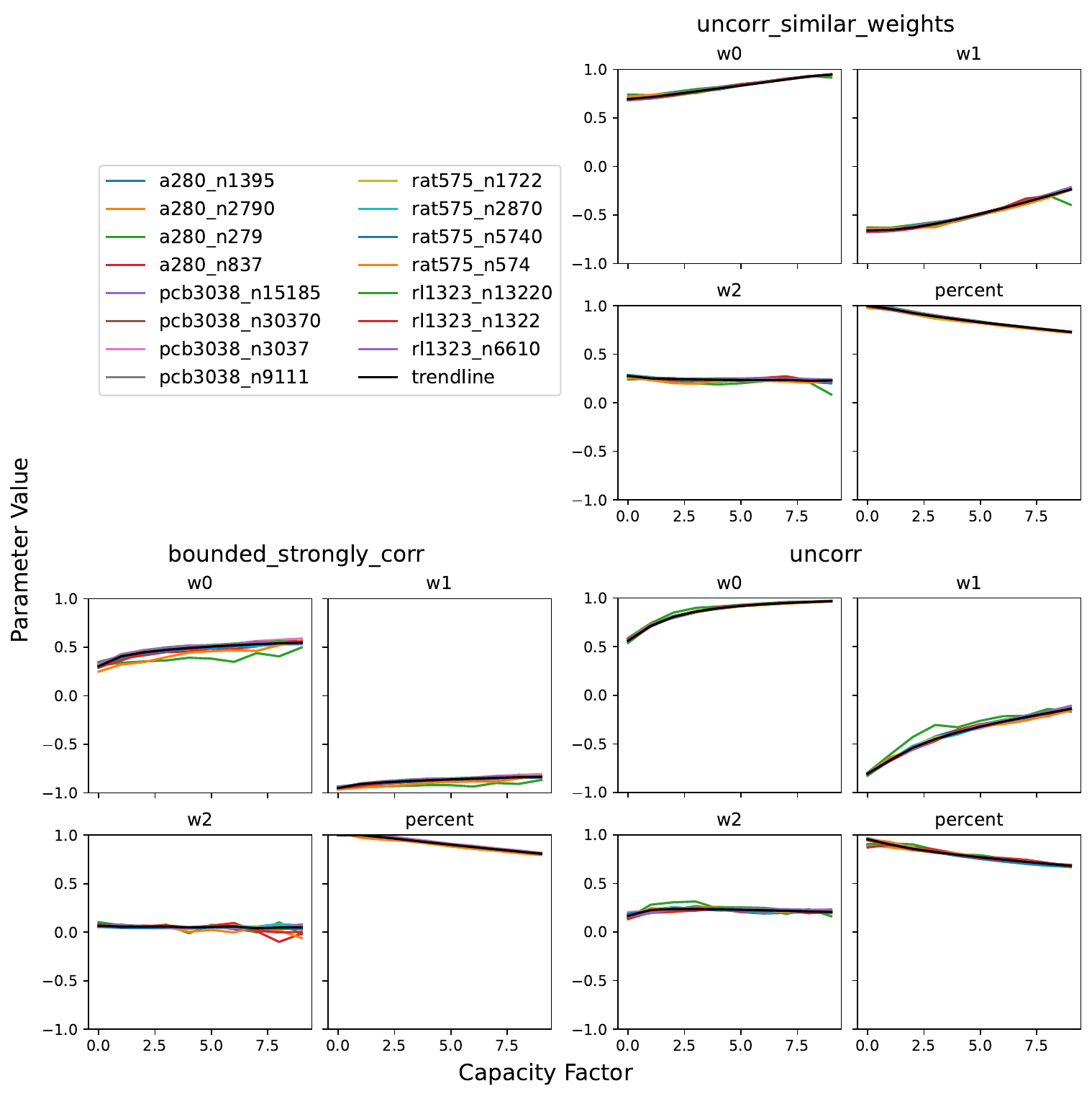}
        \caption{Plot of the optimal parameter values found by our metaheuristic GA in each training instance against the capacity factor, along with a line of best fit. Parameter values were normalized such that $w_0^2+w_1^2+w_2^2=1$.}
        
        \label{fig:weights_vs_x7}
\end{figure}

From figure \ref{fig:weights_vs_x7}, we can see that the optimal parameter values are almost entirely dependent on the capacity factor, and that our evolved solutions are a good fit.

\subsection{Heuristic Design}
By using our SR evolved solutions to predict the optimal value of each parameter in an individual's genotype, we can speed up our metaheuristic algorithm to the point that it becomes a competitive packing initialization heuristic. Our heuristic design is as follows: for each feature set {\tt 3T}...{\tt 6T}, we predict the value of each parameter $w_0$...$w_n$,$p$ using our SR solutions. For each individual with genotype $[w_0,...,w_n,p]$, we sort the items by their score function, the dot product of the weight vector $[w_0,\cdots,w_n]$ and the feature vector $[f_0,\cdots,f_n]$ from the feature set. For example, when using the {\tt 4T} feature set, this feature vector would be $[x_0,x_1,x_0*x_1]$, where $x_0$ and $x_1$ are the standardized {\tt IPR} and {\tt rDist} respectively. We then pack items in decreasing order of their score function until the objective value reaches a maximum. However, in contrast to algorithms like packIterative\cite{packiterative}, which have to start from zero packed items, we already have a good guess of the optimal knapsack weight. Therefore, we start by packing items until reaching a capacity of $pW$. We then search for the optimal packing in both directions (more items packed vs less items packed) as follows: we first take 1 step forward, packing/unpacking a single item depending on the direction. As long as the objective value increases, we double the step size. However, if the objective value decreases, we revert the changes, unpacking/packing the items, and decrease the step size by a factor of 8. The constants 2 and 8, as well as the initial step size of 1 were chosen through a preliminary hyperparameter search to minimize the number of costly objective value calculations performed by our heuristic. 

After determining the maximum objective value for the individual, we mutate its weights $w_0,...,w_n$ in a (1+1) EA with gaussian noise sampled from $\mathcal{N}(0,0.5)$. Again, the mutation strength was chosen through a preliminary search to achieve the highest objective value in the fewest generations. Instead of mutating the percent parameter $p$, we keep a running estimate of $p$ based on the optimal knapsack weights of evaluated individuals. To stabilize estimates while still adapting quickly, we test whether the new sample $p_i$ differs from the estimate $\hat{p}$ by more than 10\%. If so, then we take the average of the two for more stability $\hat{p}\leftarrow\frac{p_i+\hat{p}}{2}$. If not, then we simply replace the old estimate with the new sample $\hat{p}\leftarrow p_i$. This scheme was also found to reduce the number of objective value calculations in preliminary experiments.

\section{Experiments and Results}
Having designed a family of initialization heuristics, one for each feature set {\tt 3T}...{\tt 6T}, we now compare the heuristics both against each other and against previous SOTA heuristics. For these experiments, we choose 5 previously unseen TSP instances of roughly log-spaced size, including both the smallest and largest instances: {\tt eil51}, {\tt lin318}, {\tt d2103}, {\tt usa13059}, and {\tt pla85900}. As before, we take 90 TTP instances from each category, with capacity factors $C\in[1...10]$, KP type in {\tt bounded\_strongly\_corr}, {\tt uncorr\_similar\_weights}, and {\tt uncorr}, and item factor $F\in\{1,5,10\}$ for a total of $450$ TTP instances. We report both the final objective value obtained by the heuristic and the number of objective value calculations, which is a proxy for the computational cost that is independent of programming language and computer hardware. We first compare our heuristics against each other to determine the best one, and then rerun the best heuristic on all benchmark instances and compare to previous initialization schemes.

\subsection{Internal Comparisons} 
\begin{figure}
     \centering
     \includegraphics[width=\textwidth]{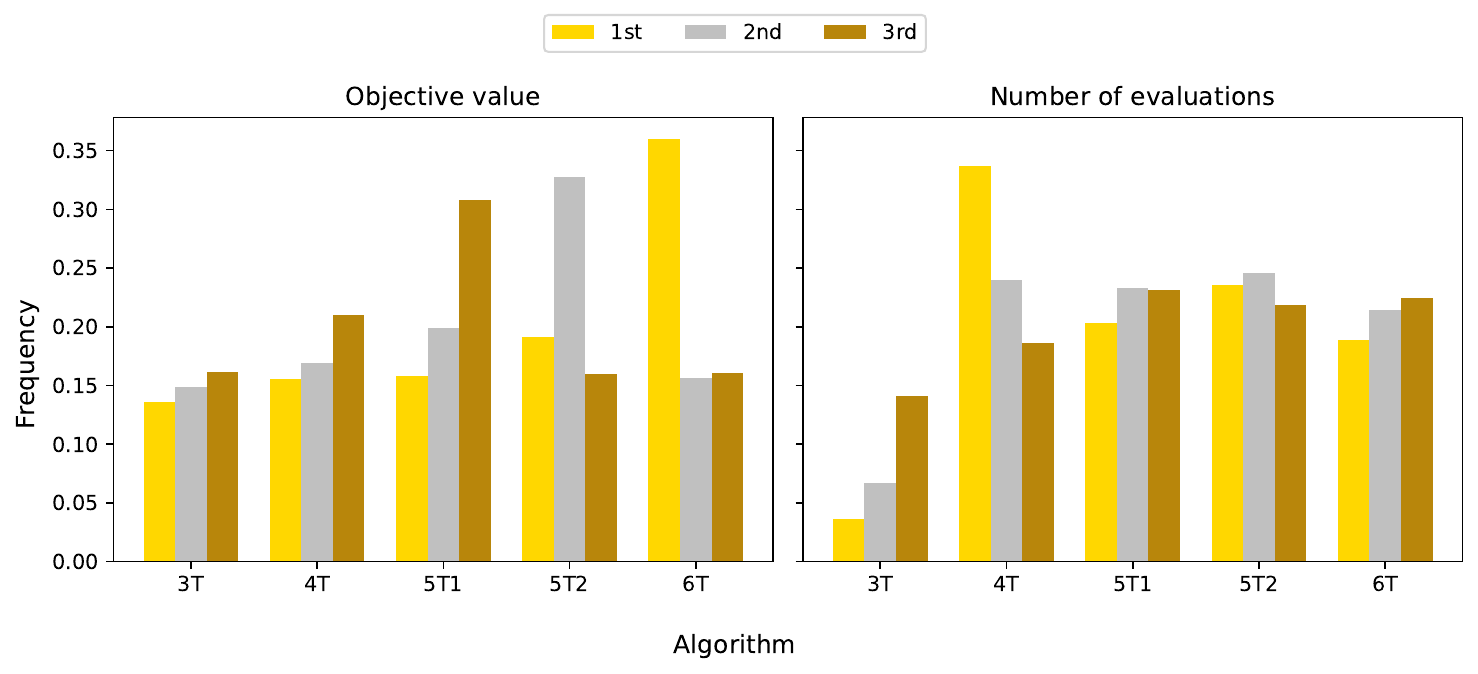}
        \caption{Ranking results for all heuristics on all 450 instances. The {\tt 6T} heuristic is has the best objective most frequently, but the {\tt 4T} algorithm generally requires fewer objective value computations.}
        \label{fig:intra_heuristic}
\end{figure}

In this section we compare our proposed heuristics against each other to determine which one performs the best. We evaluate each heuristic algorithm 30 times on each TTP instance, and each time we rank the algorithms in terms of their objective values and number of evaluations. For an equal comparison, we share the same starting tour between all heuristics. The results are shown in figure \ref{fig:intra_heuristic}

The results show that the {\tt 6T} heuristic has the highest frequency of best objective value, but the {\tt 4T} heuristic often requires the fewest number of evaluations. However, as the computational cost of the later optimization schemes often outweighs the cost of the initialization heuristic in practical TTP solvers, we choose the {\tt 6T} heuristic to compare against SOTA heuristics.

\subsection{External Comparisons} 
\begin{figure}
     \centering
     \includegraphics[width=\textwidth]{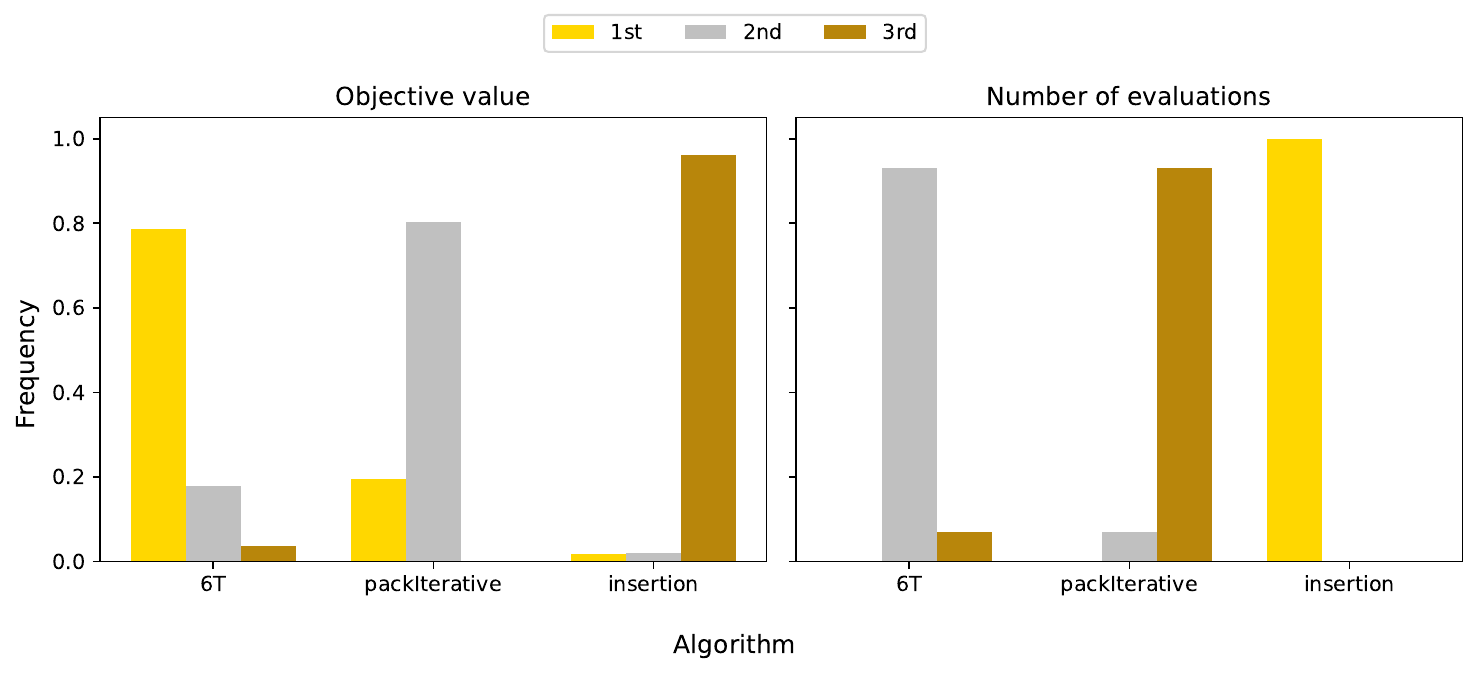}
        \caption{Ranking results for all heuristics on all 450 instances. {\tt 6T} is often the best in terms of objective value, and generally requires fewer objective value computations than {\tt packIterative}, but is slower than {\tt Insertion}.}
        
        \label{fig:inter_heuristic}
\end{figure}

In this section we compare the {\tt 6T} heuristic against initialization heuristics from the literature. Specifically, since {\tt Insertion}\cite{matls} is an improvement upon the similar {\tt DH}\cite{dh} and {\tt SH}\cite{ttpbenchmarks} heuristics, we compare against {\tt Insertion} and \\{\tt packIterative}\cite{packiterative} . As before, we run our heuristic algorithm 30 times on each TTP instance and share the same tour between all algorithms. Since the other three heuristics are deterministic, they are run only once to determine their performance across all 30 trials. At each trial we rank the algorithms in terms of the objective value and number of evaluations. The results are shown in figure \ref{fig:inter_heuristic}

\section{Conclusion and Future Work}

In this work, we presented a set of 5 novel packing initialization heuristics for the traveling thief problem. Rather than being designed by human intuition, these heuristics were derived from the interpretable modeling of near-optimal packing plans using symbolic regression. Experimental results on a large range of problem instances validate the effectiveness of our approach, and we demonstrate improved objective values compared to previous SOTA initialization heuristics. While we have focused on comparisons between heuristics in this paper, we do not anticipate that the best algorithms will utilize only a single initialization heuristic. In fact, even though our heuristic almost always achieves a higher objective value than the {\tt Insertion} heuristic, {\tt Insertion} has such a low cost that in practice it makes sense to always run {\tt Insertion} in addition to our heuristic, similarly to Majid et al.\cite{coco}, on the off chance that {\tt Insertion} will perform better. In a more complicated scheme, we could also learn to predict which of the proposed heuristics will perform the best on a given TTP instance from various features of the instance, similarly to Wagner et al.\cite{ttpalgorithmbench}. 

In addition, while we have evolved separate functions to predict the optimal parameter values for each KP type, we hypothesize that more general functions can be evolved that use more fundamental properties of the TTP instances, for example, the covariance matrix of the joint distribution of weights and profits. Therefore, future work could expand the current benchmark set of TTP instances to include other distributions of weights and profits, and use this additional training data to evolve a single function for each parameter that will work across all TTP instances.

\section{Acknowledgements}

This work was performed in part using high-performance computing equipment at Amherst College obtained under National Science Foundation Grant No. 2117377. Any opinions, findings, and conclusions or recommendations expressed in this publication are those of the authors and do not necessarily reflect the views of the National Science Foundation. The authors would like to thank Scott Kaplan, Ryan Boldi, Bill Tozier, Tom Helmuth, Edward Pantridge and other members of the PUSH lab for their insightful comments and suggestions.

\newpage
\bibliographystyle{splncs04}
\bibliography{refs}


\section{Frequency of Input Variables in Evolved Solutions}
In this section we present the rest of the results of the experiment described in section 3.2. Shown below are the occurrence frequencies of each input variable in functions evolved to predict the optimal value of a parameter in a metaheuristic GA. The conclusions are the same across the board: the $x_7$ (capacity factor) variable is essential, and the other input variables are irrelevant, except for maybe the $x_4$ (item factor) variable, which is occasionally present in a large fraction of solutions.

\begin{table}
\caption{Percent of evolved solutions containing each input variable for each parameter in the genotype of an individual from the {\tt 3T} metaheuristic GA.}\label{ftable1}
\begin{tabular}{|c||c|c|c|c|c||c|c|c|c|c||c|c|c|c|c|}

\hline
     & \multicolumn{5}{c||}{bounded-strongly-corr}& \multicolumn{5}{c||}{uncorr-similar-weights} &\multicolumn{5}{c|}{uncorr}\\\hline
     & $x_1$ & $x_2$ & $x_3$ & $x_4$& $x_7$&  $x_1$ & $x_2$ & $x_3$ & $x_4$&$x_7$&  $x_1$ & $x_2$ & $x_3$ & $x_4$& $x_7$ \\\hline
$w0$ & 0.4 & 0.3 & 0.17 & 0.4 & {\bf 1.0} & 0.1 & 0.47 & 0.07 & 0.3 & {\bf 1.0} & 0.02 & 0.05 & 0.18 & 0.05 & {\bf 1.0}\\
$w1$ & 0.27 & 0.45 & 0.03 & 0.38 & {\bf 1.0} & 0.18 & 0.33 & 0.1 & 0.28 & {\bf 1.0} & 0.15 & 0.13 & 0.18 & 0.05 & {\bf 1.0}\\
percent & 0.03 & 0.07 & 0.0 & 0.05 & {\bf 1.0} & 0.1 & 0.12 & 0.18 & 0.08 & {\bf 0.93} & 0.28 & 0.12 & 0.07 & 0.2 & {\bf 1.0}\\\hline
\end{tabular}
\end{table}

\begin{table}
\caption{Percent of evolved solutions containing each input variable for each parameter in the genotype of an individual from the {\tt 4T} metaheuristic GA. These results were also shown in section 3.3 of the paper}\label{ftable1}
\begin{tabular}{|c||c|c|c|c|c||c|c|c|c|c||c|c|c|c|c|}

\hline
     & \multicolumn{5}{c||}{bounded-strongly-corr}& \multicolumn{5}{c||}{uncorr-similar-weights} &\multicolumn{5}{c|}{uncorr}\\\hline
     & $x_1$ & $x_2$ & $x_3$ & $x_4$& $x_7$&  $x_1$ & $x_2$ & $x_3$ & $x_4$&$x_7$&  $x_1$ & $x_2$ & $x_3$ & $x_4$& $x_7$ \\\hline
$w0$ & 0.17 & 0.37 & 0.07 & 0.4 & {\bf 1.0} & 0.27 & 0.1 & 0.17 & 0.17 & {\bf 1.0} & 0.2 & 0.13 & 0.07 & 0.03 & {\bf 0.97}\\
$w1$ & 0.37 & 0.17 & 0.03 & 0.43 & {\bf 1.0} & 0.1 & 0.0 & 0.0 & 0.1 & {\bf 1.0} & 0.07 & 0.0 & 0.18 & 0.0 & {\bf 1.0}\\
$w2$ & 0.1 & 0.17 & 0.03 & 0.9 & {\bf 1.0} & 0.57 & 0.5 & 0.07 & 0.6 & {\bf 1.0} & 0.4 & 0.1 & 0.07 & 0.5 & {\bf 0.93}\\
percent & 0.1 & 0.1 & 0.1 & 0.13 & {\bf 1.0} & 0.0 & 0.0 & 0.1 & 0.03 & {\bf 1.0} & 0.03 & 0.03 & 0.0 & 0.0 & {\bf 1.0}\\\hline
\end{tabular}
\end{table}

\begin{table}
\caption{Percent of evolved solutions containing each input variable for each parameter in the genotype of an individual from the {\tt 5T1} metaheuristic GA.}\label{ftable1}
\begin{tabular}{|c||c|c|c|c|c||c|c|c|c|c||c|c|c|c|c|}

\hline
     & \multicolumn{5}{c||}{bounded-strongly-corr}& \multicolumn{5}{c||}{uncorr-similar-weights} &\multicolumn{5}{c|}{uncorr}\\\hline
     & $x_1$ & $x_2$ & $x_3$ & $x_4$& $x_7$&  $x_1$ & $x_2$ & $x_3$ & $x_4$&$x_7$&  $x_1$ & $x_2$ & $x_3$ & $x_4$& $x_7$ \\\hline
$w0$ & 0.2 & 0.57 & 0.0 & 0.53 & {\bf 1.0} & 0.1 & 0.0 & 0.0 & 0.13 & {\bf 1.0} & 0.1 & 0.1 & 0.1 & 0.3 & {\bf 1.0}\\
$w1$ & 0.17 & 0.5 & 0.07 & 0.5 & {\bf 1.0} & 0.0 & 0.03 & 0.17 & 0.17 & {\bf 1.0} & 0.13 & 0.1 & 0.07 & 0.2 & {\bf 1.0}\\
$w2$ & 0.63 & 0.8 & 0.1 & 0.77 & {\bf 1.0} & 0.17 & 0.43 & 0.17 & 0.57 & {\bf 0.9} & 0.47 & 0.8 & 0.13 & 0.7 & {\bf 0.97}\\
$w3$ & 0.03 & 0.3 & 0.1 & 0.83 & {\bf 1.0} & 0.0 & 0.1 & 0.0 & 0.3 & {\bf 1.0} & 0.17 & 0.3 & 0.13 & 0.53 & {\bf 1.0}\\
percent & 0.37 & 0.1 & 0.2 & 0.17 & {\bf 0.9} & 0.07 & 0.11 & 0.07 & 0.04 & {\bf 0.93} & 0.07 & 0.03 & 0.0 & 0.24 & {\bf 1.0}\\\hline
\end{tabular}
\end{table}

\begin{table}
\caption{Percent of evolved solutions containing each input variable for each parameter in the genotype of an individual from the {\tt 5T2} metaheuristic GA.}\label{ftable1}
\begin{tabular}{|c||c|c|c|c|c||c|c|c|c|c||c|c|c|c|c|}

\hline
     & \multicolumn{5}{c||}{bounded-strongly-corr}& \multicolumn{5}{c||}{uncorr-similar-weights} &\multicolumn{5}{c|}{uncorr}\\\hline
     & $x_1$ & $x_2$ & $x_3$ & $x_4$& $x_7$&  $x_1$ & $x_2$ & $x_3$ & $x_4$&$x_7$&  $x_1$ & $x_2$ & $x_3$ & $x_4$& $x_7$ \\\hline
$w0$ & 0.53 & 0.33 & 0.13 & 0.33 & {\bf 1.0} & 0.03 & 0.07 & 0.1 & 0.13 & {\bf 1.0} & 0.03 & 0.1 & 0.0 & 0.13 & {\bf 1.0}\\
$w1$ & 0.57 & 0.53 & 0.3 & 0.43 & {\bf 1.0} & 0.03 & 0.03 & 0.1 & 0.1 & {\bf 1.0} & 0.23 & 0.07 & 0.33 & 0.17 & {\bf 0.9}\\
$w2$ & 0.4 & 0.4 & 0.07 & 0.77 & {\bf 1.0} & 0.17 & 0.83 & 0.07 & 0.8 & {\bf 0.97} & 0.2 & 0.07 & 0.13 & 0.03 & {\bf 0.97}\\
$w3$ & 0.5 & 0.67 & 0.1 & 0.47 & {\bf 1.0} & 0.1 & 0.77 & 0.0 & 0.73 & {\bf 1.0} & 0.5 & 0.1 & 0.2 & 0.4 & {\bf 1.0}\\
percent & 0.23 & 0.13 & 0.03 & 0.1 & {\bf 0.97} & 0.33 & 0.03 & 0.07 & 0.33 & {\bf 1.0} & 0.33 & 0.03 & 0.03 & 0.3 & {\bf 1.0}\\\hline
\end{tabular}
\end{table}

\begin{table}
\caption{Percent of evolved solutions containing each input variable for each parameter in the genotype of an individual from the {\tt 6T} metaheuristic GA.}\label{ftable1}
\begin{tabular}{|c||c|c|c|c|c||c|c|c|c|c||c|c|c|c|c|}

\hline
     & \multicolumn{5}{c||}{bounded-strongly-corr}& \multicolumn{5}{c||}{uncorr-similar-weights} &\multicolumn{5}{c|}{uncorr}\\\hline
     & $x_1$ & $x_2$ & $x_3$ & $x_4$& $x_7$&  $x_1$ & $x_2$ & $x_3$ & $x_4$&$x_7$&  $x_1$ & $x_2$ & $x_3$ & $x_4$& $x_7$ \\\hline
$w0$ & 0.23 & 0.4 & 0.07 & 0.57 & {\bf 1.0} & 0.13 & 0.03 & 0.1 & 0.13 & {\bf 1.0} & 0.0 & 0.23 & 0.1 & 0.03 & {\bf 1.0}\\
$w1$ & 0.1 & 0.4 & 0.03 & 0.3 & {\bf 1.0} & 0.0 & 0.03 & 0.1 & 0.13 & {\bf 1.0} & 0.1 & 0.1 & 0.07 & 0.17 & {\bf 1.0}\\
$w2$ & 0.37 & 0.07 & 0.1 & 0.13 & {\bf 1.0} & 0.23 & 0.3 & 0.07 & 0.27 & {\bf 1.0} & 0.17 & 0.9 & 0.1 & 0.87 & {\bf 1.0}\\
$w3$ & 0.13 & 0.17 & 0.07 & 0.43 & {\bf 1.0} & 0.13 & 0.57 & 0.03 & 0.3 & {\bf 1.0} & 0.1 & 0.93 & 0.2 & 0.83 & {\bf 0.97}\\
$w4$ & 0.07 & 0.27 & 0.0 & 0.37 & {\bf 1.0} & 0.33 & 0.43 & 0.0 & 0.4 &{\bf 0.97 }& 0.17 & 0.8 & 0.0 & 0.83 & {\bf 0.97}\\
percent & 0.03 & 0.0 & 0.1 & 0.0 & {\bf 1.0} & 0.03 & 0.1 & 0.03 & 0.07 & {\bf 0.97} & 0.2 & 0.03 & 0.13 & 0.47 & {\bf 1.0}\\\hline
\end{tabular}
\end{table}

\section{Optimal Parameter Value Plots}

In this section we display the plots of the optimal parameter values found by our metaheuristic genetic algorithm against the capacity factor of each TTP instance. In each case, the weights were normalized so that the sum of their squares equals 1. In each case, the optimal parameter value depends almost entirely on the capacity factor, and can often be approximated with a simple linear function with the notable exception of $w_0$ in problems of the {\tt uncorr} knapsack type. In addition, as the number of terms increases from the {\tt 2T} to the {\tt 6T} algorithm, the optimal values of the weights becomes more noisy due to the larger dimensionality and the higher complexity.

\begin{figure}
     \centering
     \includegraphics[width=\textwidth]{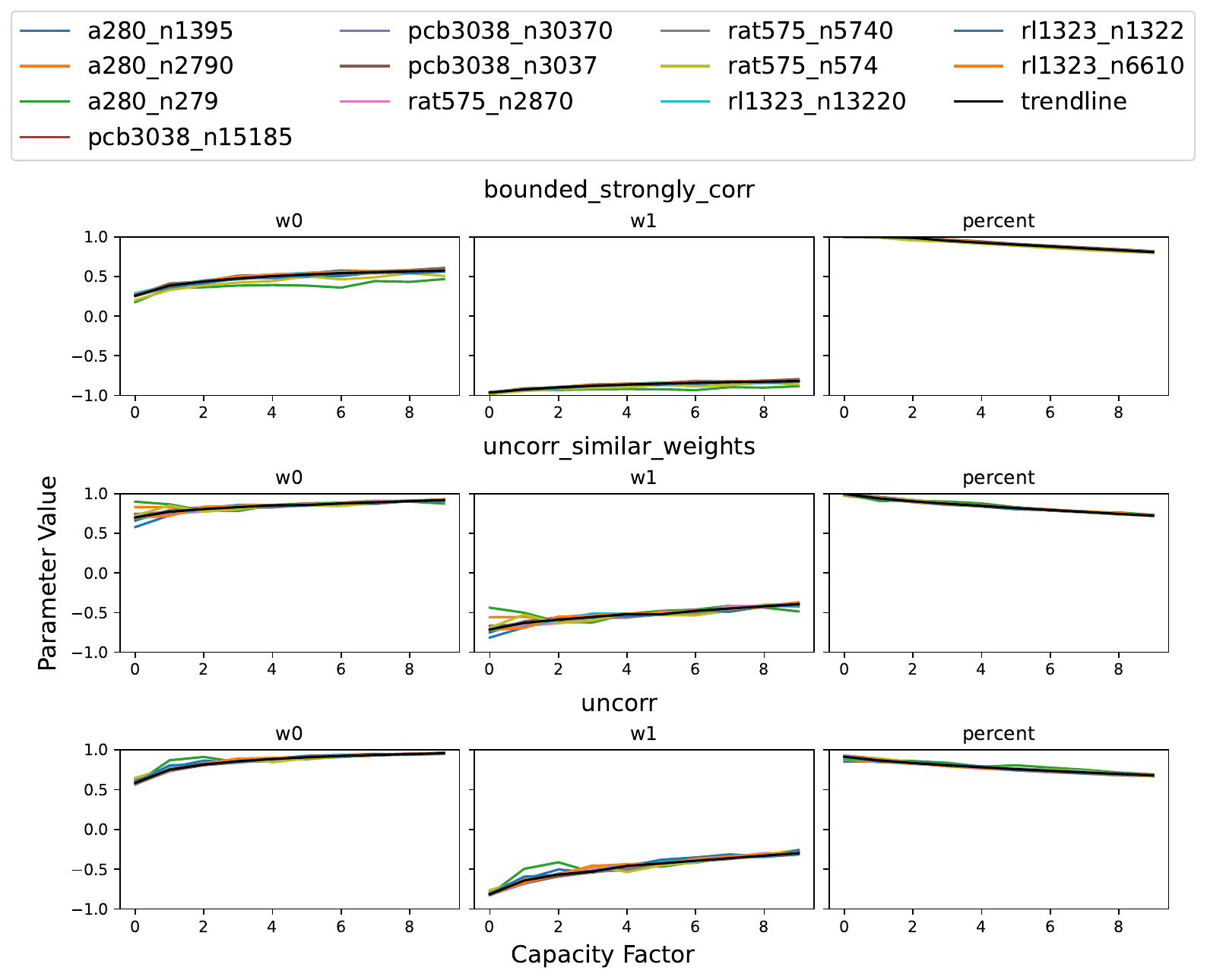}
        \caption{Plot of the optimal parameter values against the capacity factor for our {\tt 2T} metaheuristic algorithm. The solution evolved by SR is plotted in black.}
        \label{fig:2T}
\end{figure}

\begin{figure}
     \centering
     \includegraphics[width=\textwidth]{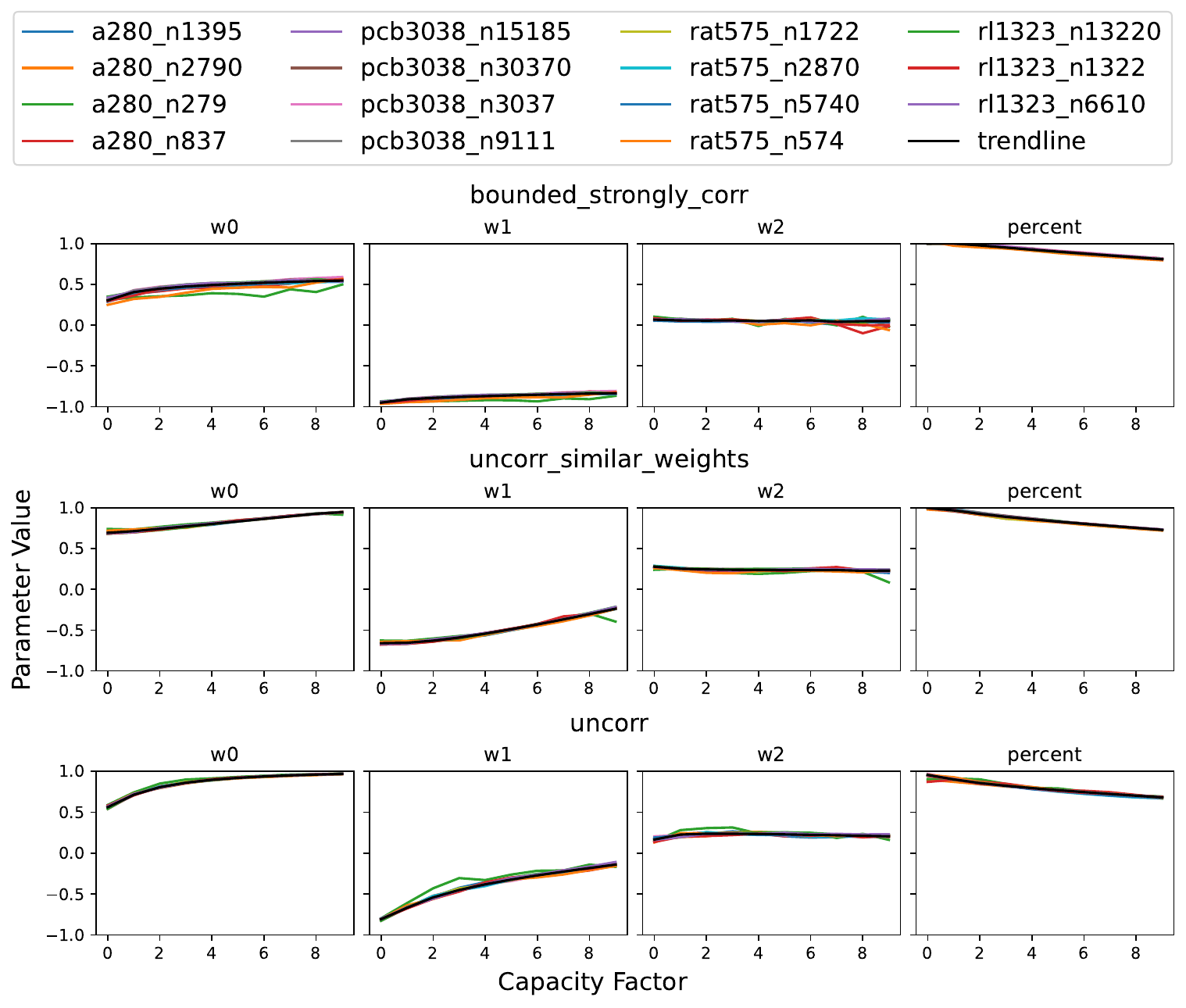}
        \caption{Plot of the optimal parameter values against the capacity factor for our {\tt 3T} metaheuristic algorithm. The same plot is shown in a different format in section 3.2 of the paper. The solution evolved by SR is plotted in black.}
        \label{fig:3T}
\end{figure}

\begin{figure}
     \centering
     \includegraphics[width=\textwidth]{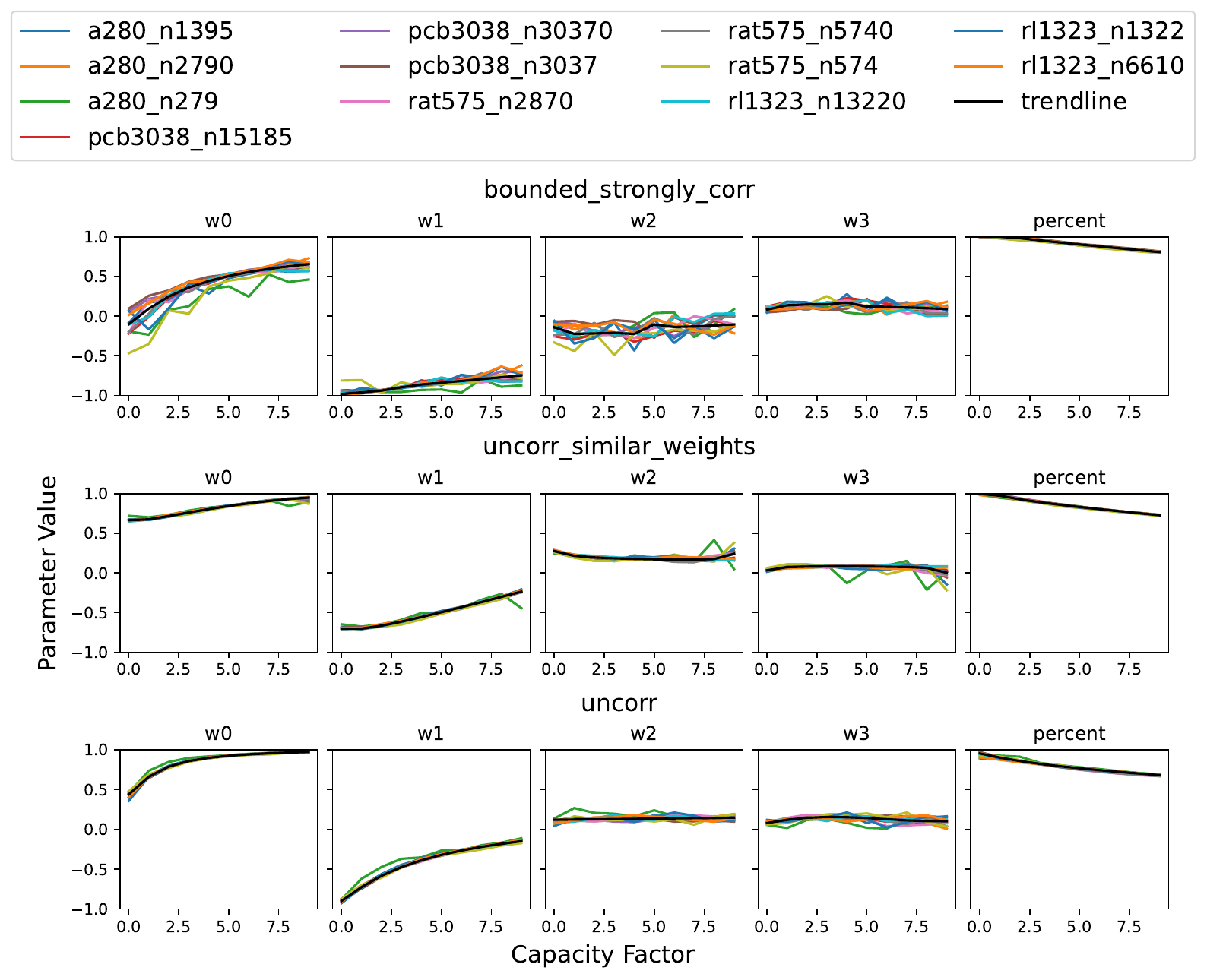}
        \caption{Plot of the optimal parameter values against the capacity factor for our {\tt 5T1} metaheuristic algorithm. The solution evolved by SR is plotted in black.}
        \label{fig:5T1}
\end{figure}

\begin{figure}
     \centering
     \includegraphics[width=\textwidth]{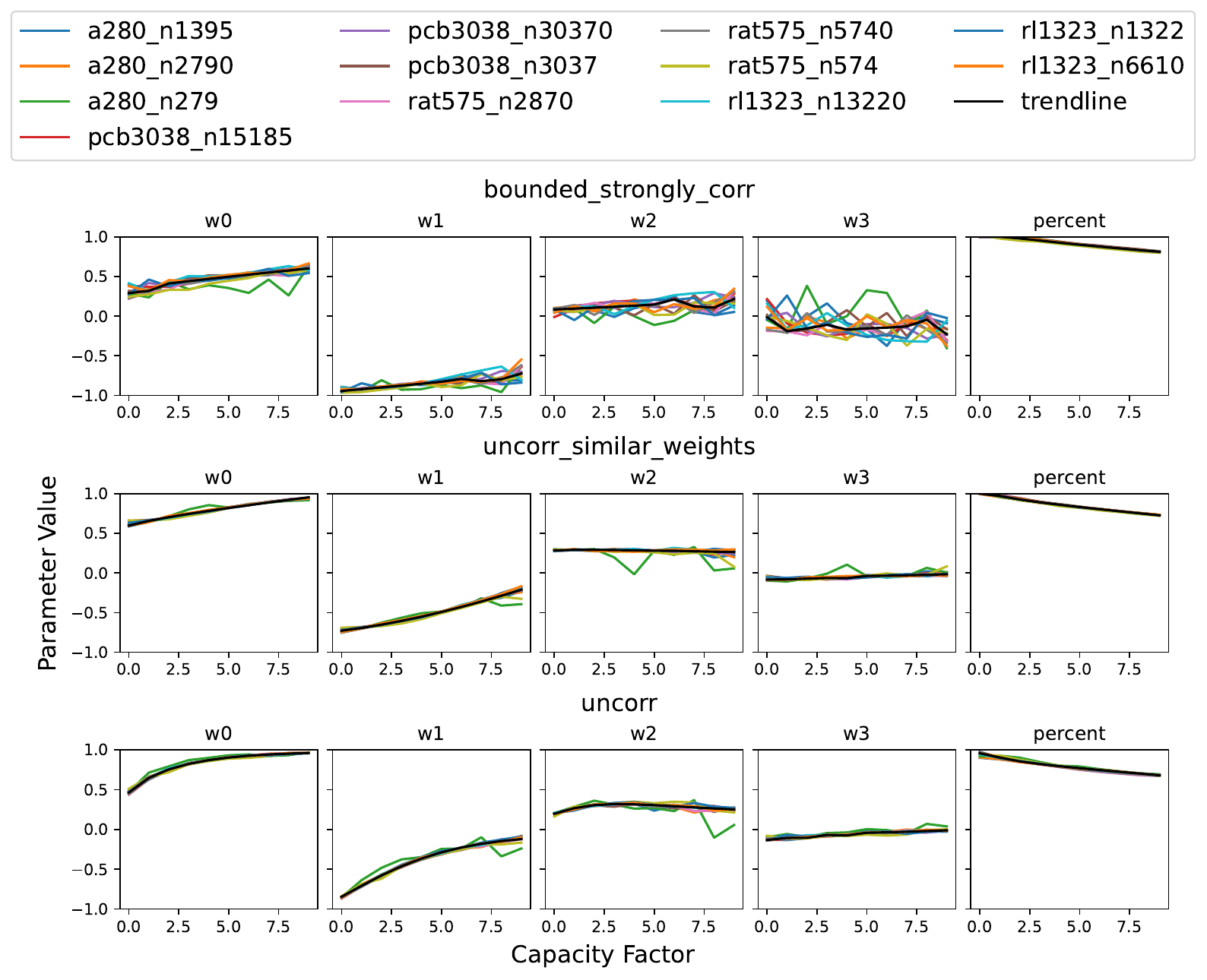}
        \caption{Plot of the optimal parameter values against the capacity factor for our {\tt 5T2} metaheuristic algorithm. The solution evolved by SR is plotted in black.}
        \label{fig:5T2}
\end{figure}

\begin{figure}
     \centering
     \includegraphics[width=\textwidth]{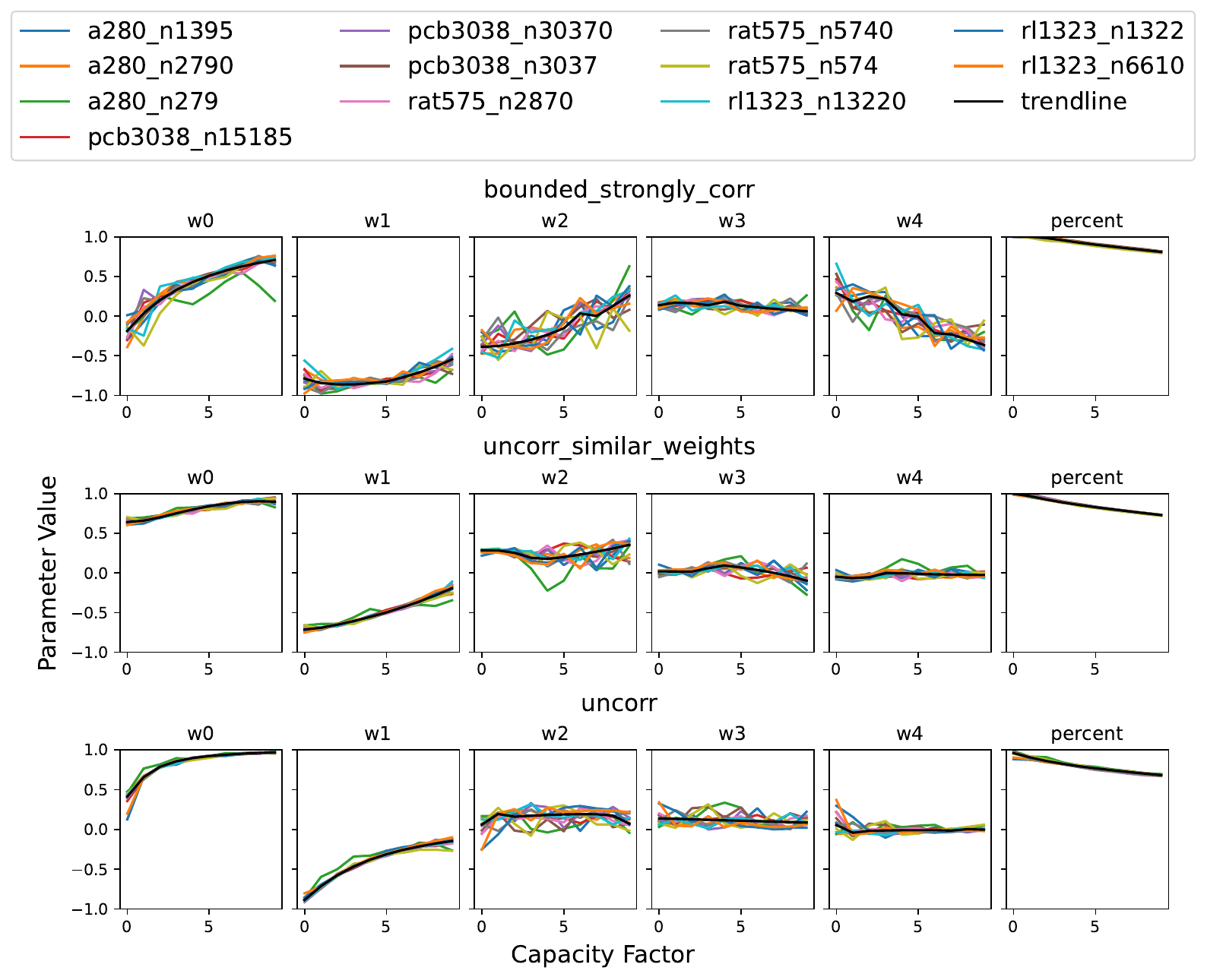}
        \caption{Plot of the optimal parameter values against the capacity factor for our {\tt 6T} metaheuristic algorithm. The solution evolved by SR is plotted in black.}
        \label{fig:6T}
\end{figure}

\end{document}